\documentclass[11pt,a4paper]{article}
\usepackage[hyperref]{acl2021}
\usepackage{times}
\usepackage{latexsym}

\usepackage{microtype}
\usepackage{booktabs}
\usepackage{amsmath,amssymb,amsthm,amscd,epsfig,amsfonts,rotating,bm,bbm}
\usepackage{multirow}

\aclfinalcopy 
\def\aclpaperid{2747} 

\newcommand{\todo}[1]{\textcolor{blue}{}}

\newcommand{\runzhe}[1]{\textcolor{cyan}{}}
\newcommand{\jingxiao}[1]{\textcolor{red}{}}

\title{Improving Dialog Systems for Negotiation with Personality Modeling}

\author{Runzhe Yang\footnote[1]{} \\ Princeton University \\  \texttt{runzhey@princeton.edu}   \\
\And Jingxiao Chen\thanks{~Authors contributed equally.} \\Shanghai Jiao Tong University \\ \texttt{timemachine@sjtu.edu.cn} \\
\And Karthik Narasimhan \\ Princeton University \\  \texttt{karthikn@princeton.edu} }

\date{}

\begin{document}
\maketitle
\begin{abstract}
    In this paper, we explore the ability to model and infer personality types of opponents, predict their responses, and use this information to adapt a dialog agent's high-level strategy in negotiation tasks. Inspired by the idea of incorporating a theory of mind (ToM) into machines, we introduce a probabilistic formulation to encapsulate the opponent's personality type during both learning and inference. We test our approach on the {\sc CraigslistBargain} dataset~\cite{he2018emnlp} and show that our method using ToM inference achieves a 20\% higher dialog agreement rate compared to baselines on a mixed population of opponents. We also find that our model displays diverse negotiation behavior with different types of opponents.\footnote{Code and data available at \url{https://github.com/princeton-nlp/NegotiationToM} }
    \end{abstract}
\section{Introduction}
\label{sec:introduction}

Developing dialog systems for negotiation is challenging since the task requires a combination of good communication skills and strategic reasoning capabilities~\cite{traum2008multi,young2013ieee,keizer2017eacl}.
While recent neural models \cite{wen2017ecal, dhingra2017acl,zhou2019sigdial, he2018emnlp} have shown that useful dialogue strategies can be learned from offline corpora, they do not explicitly model the mental state of other agents, which can make it challenging to generate tailored strategies and utterances for different types of opponents. 

In this paper, we introduce a new framework for generating strategic dialog inspired by the idea of Theory of Mind (ToM)
from cognitive science~\cite{premack1978does, bruner1981intention, wimmer1983beliefs}. When negotiating with others, humans innately infer the intention of the other party, and guess how their own utterances would affect the opponent's mental state. To emulate this capability in machines, we train a first-order ToM model to predict an opponent's response given the current state and the agent's own possible utterances. This first-order ToM model can then be incorporated into dialog agents to enable one-step lookaheads during inference. 

In order to predict future responses, we model the opponent's personality type as a intermediate variable ($z$), which can be predicted using the dialogue history. We use this predicted personality, along with the previous state and utterance to calculate the likelihood of the opponent's next state for all possible actions that our agent can take in the current state. This allows us to compute an expected value of return for each action, which is subsequently used to produce a policy for our agent. 
We propose two variants of our ToM-based dialog agent -- an \textbf{explicit} version that outputs the opponent type as an intermediate prediction, and an \textbf{implicit} version that models the opponent type as a latent variable. Both models can be instantiated as end-to-end neural networks and can be trained using reinforcement learning. 

Our approach differs from existing opponent modeling work \cite{lee2018answerer, hadjinikolis2013opponent, oren2009arguing, rienstra2013opponent, HeB16} 
in three aspects: 1) it provides strategic benefit during \textit{inference} which leads to more successful negotiations, 2) it can flexibly adjust the degree of dependence on ToM predictions by changing a temperature parameter, and 3) it utilizes text utterances to infer types of opponents, thereby capturing side information ({\em e.g.}, emotion) that is useful yet absent from standard dialog state transitions.

We perform experiments on a modified version of the {\sc CraigslistBargain} negotiation task~\cite{he2018emnlp}, where the agent is matched with different opponents from diverse populations (e.g., cooperative, competitive, and aggressive negotiators), without being provided information about their identity.  
Empirically, our method outperforms several baselines on the task by completing more deals and achieving higher utility. For instance, our model achieves about 20\% higher dialog agreement rate and utility than a baseline dialog manager trained with reinforcement learning. Our analysis reveals that the agent demonstrates diverse negotiation behavior and adapts well to different types of opponents. 



\section{Related Work}
\label{sec:related}
\begin{figure*}[tbp]
    \centering
    \includegraphics[width=0.95\textwidth]{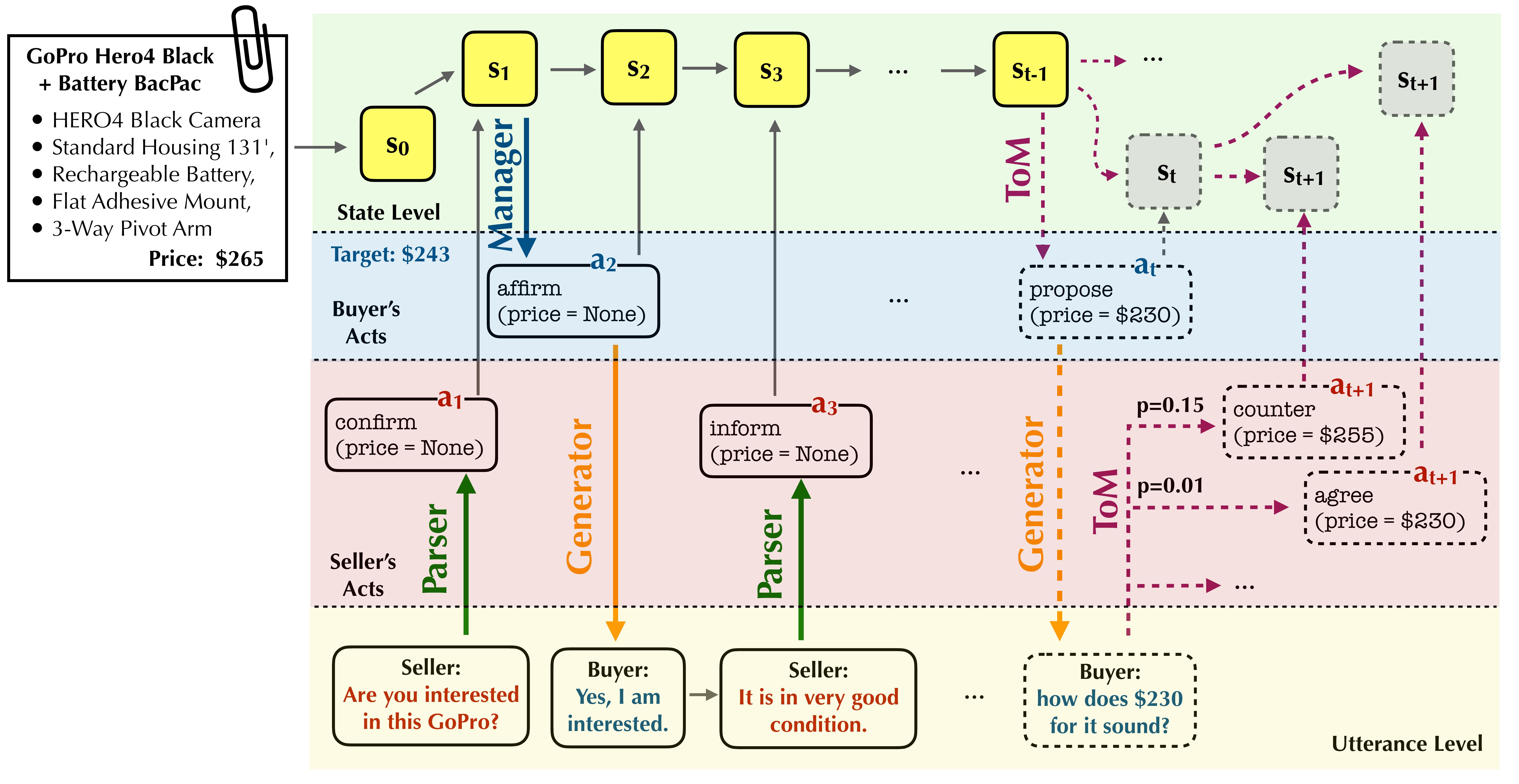}
    \vspace{-0.5em}
    \caption{Our Theory of Mind (ToM) framework of negotiation systems. The interaction between a {\em buyer} and a {\em seller} can be divided into three levels: The utterance level, dialog act level, and state level. The parser extracts an intent and key information ({\em e.g.}, price) from an input utterance as a dialog act. Both intents and key information, along with the context ({{\em e.g.} description about the item}), contribute to the state of dialog. The traditional RL-based dialog manager decides a dialog act based on the current state. And the generator converts the abstract dialog act back to a natural language utterance, also based on the previous state. The first-order ToM model explicitly predicts the response of the opponent and the state transition, which supports more strategic negotiation.}
    \vspace{-0.5em}
    \label{fig:framework}
\end{figure*}

\paragraph{Speaker-follower models and rational speech acts.}
Our work is related to recent papers using the Rational Speech Acts (RSA) model for natural  language~\cite{goodman2013knowledge, monroe2015learning, goodman2016pragmatic, shen2019pragmatically}.
RSA has also been applied to language grounding \cite{andreas2016reasoning} and vision-language navigation \cite{fried2018speaker}. Our first-order theory of mind modeling is different since we learn how the speaker's intent and utterance affect the opponent's reaction, instead of assuming the optimality of the listener in the speaker's mind. Recent RSA model \cite{white2020} considers speakers and listeners in resource-constrained settings, while we do not enforce constraints on opponents.


Our approach with explicit characteristic modeling is also similar to the ToMnet \cite{rabinowitz2018icml}, which uses a multi-agent reinforcement learning setting to learn identity embeddings of populations from past trajectories, and predict the mental state of an agent using the current trajectory. However, our first-order ToM models for negotiation also take utterances into account, which makes improving upon a base RL policy non-trivial.

\paragraph{Theory of Mind in dialog systems.} Theory of mind for modeling user personality types and predicting responses has been studied in the context of building user simulators~\cite{GeorgilaHL06, RieserL06} for training RL-based dialog systems, and to make dialog systems explainable~\cite{ChandrasekaranY17}.  Recent work on dialog policy learning has employed theory of mind with a focus on specific domains. The Recursive Mental Model (RMM) \cite{Roman2020} was proposed for navigation settings, where questions and answers are generated between a navigating agent and a guiding agent. Another approach --  Answerer in Questioner's Mind (AQM) \cite{lee2018answerer} --  tackled an answer guessing game with information-theoretic methods. In these domains, the opponents are assumed to be cooperative, while our method is applicable for interacting with both cooperative and competitive opponents. \citet{wang-etal-2019-persuasion} designed a personalized dialog system based on sentence-level strategy annotations to encourage donation to a specific charity. Recently, \citet{JangLK20} employed Bayesian-optimal Monte-Carlo planning for end-to-end dialog generation at the utterance level. However, their method only models the latent goal of the opponent instead of potential responses like we do.

\paragraph{Opponent modeling in RL.} Apart from dialog systems, opponent modeling has been explored in other multi-agent reinforcement learning settings \cite{WenYLWP19, OstenKM17, HeB16, hadjinikolis2013opponent, rienstra2013opponent}. Our approach differs from these works by: 1) providing strategic benefit during real-time inference, 2) adjusting the degree of dependence on the ToM predictions through a temperature parameter, and 3) utilizing text utterances in the dialog to infer types of opponents, thereby capturing side information that is useful yet absent from standard state transitions.

\vspace{-0.2em}
\section{Framework}
\label{sec:method}
\vspace{-0.2em}


\paragraph{Task.} 
We consider a task-oriented dialog setting where there are two agents, a buyer and a seller. The buyer's goal is to purchase the listed item with minimum cost, and the seller's goal is to sell the item at a price as high as possible. The item description is public for both agents, while the target prices are private for both buyer and seller. Two agents negotiate in alternating turns until they conclude with an agreement or disagreement.

\paragraph{MDP Formulation.} 
We formulate the negotiation process between two agents as a {\em  multi-agent Markov Decision Process} (MAMDP), $\langle \mathcal N, \mathcal S, \mathcal A, \mathcal P, \mathcal R, \Pi, n \rangle$. 
$\mathcal N = \{-1, 1\}$ is the set indicating two agents ({\em buyer=-1 / seller = 1}). $\mathcal A$ is the action space consisting of {\em dialog acts}. 
For example, a valid dialog act $a^i_t \in \mathcal A$ can encode the {\em intent} ({\em inform}, {\em propose}, {\em counter}, etc.) and {\em price} that the agent $i$ tries to express in the $t$-th round. 
Two agents act alternatively, i.e., if at the round $t$ only the agent $i$ moves, then at the round $t+1$ only the agent $-i$ moves.

$\mathcal S$ is the state space consisting of the negotiation status. 
We define
$s_0 \in \mathcal S$ as the initial status of the dialog, which contains the information about items to be negotiated (e.g., initial price, description). 
We also define $s_t = (s_0, a_{1}^{i}, a_{2}^{-i},  \dots,  a_{t-1}^{i}, a_{t}^{-i})$. 
In this way, the only randomness of the environment comes from the opponents policy ($s_{t-1}\rightarrow a^{-i}_{t}$), i.e.,  $s_{t-1} \rightarrow s_{t}$ is stochastic, while $(s_{t-1}, a_t^{-i}) \rightarrow s_{t}$ is deterministic.
Note that the state $s_t$ is only partially observable in reality, since one can only infer the true intent from the corresponding utterance.
We provide a summary of all the symbols used in Table \ref{tab:mdp}.

\begin{table}[t]
    \centering
    \resizebox{\columnwidth}{!}{
    \begin{tabular}{p{80pt}p{140pt}}
        \toprule
        \textbf{Symbol} & \textbf{Definition}\\
        \midrule
        $\mathcal N = \{-1, 1\}$ & Identities of the two players ({\em buyer = -1 / seller = 1}) 
        \\
		 $s_0 \in \mathcal S$ & Initial state of the dialog (e.g., list price, description).\\
 		 $a^i_t \in \mathcal A$ & dialog act of agent $i$. If $-i$, it denotes the opponent.\\
		 $s_t \in \mathcal S$ & State at the end of round $t$, $s_t:=(s_0, a_{1}^{i}, a_{2}^{-i},   \dots,  a_{t-1}^{i}, a_{t}^{-i})$ 
		 \\
		 $\mathcal P(s_t|s_{t-1})$ & Transition probability, associated with agent policies\\
		 $\pi^{i}(a^{i}_{t}|s_{t-1})$ & Probability of the agent $i$ choosing $a^i_{t}$ given the previous dialog state. \\
		 $\mathcal R = \{r^{-i}, r^{i}\}$ & Reward functions for agents $i$ and $-i$.\\
		 $n$ & Maximum length of dialogs.\\
		\midrule
		$u_t^{i}$ & Utterance of agent $i$. If $-i$, it denotes the opponent.\\
		$z^{-i}$ & Type the opponent. Annotations are available in the corpus.\\
        \bottomrule
    \end{tabular}
    }
    \vspace{-0.5em}
    \caption{Notation for our MAMDP formulation of task-oriented dialog for negotiation.}
    \label{tab:mdp}
\end{table}

\subsection{Negotiation Systems}
\label{sec:method:system}

As illustrated in Figure \ref{fig:framework}, our negotiation system encapsulates three important modules following traditional goal-oriented dialog systems \cite{young2013ieee}: 
\begin{itemize}
	\item A {\bf parser} that converts the opponent's utterance $u^{-i}_{t-1}$ to dialog act $a^{-i}_{t-1}$ (e.g., {\em ``Are you interested in this GoPro"} $\rightarrow$  \texttt{confirm(price=None)}). Since the dialog acts in our system do not intend to capture the complete semantics of a sentence, a simple rule-based parser is effective; 
	\item A {\bf manager} that decides the responding dialog act $a^i_{t}$ according to the current dialog state $s_{t-1} = (s_0, \dots, a^{-i}_{t-1})$. Our ToM model is applied to this component of the system; 
	\item A {\bf generator} that produces natural language response $u^{i}_t$ based on the current dialog act $a^{i}_t$ and the dialog state $s_{t-1}$, or equivalently $s_{t}$ (e.g., the previous dialog state + \texttt{propose(price=\$230)} $\rightarrow$ {\em ``How does \$230 for the GoPro sound?"}). It can be either deterministic to reduce computational cost or probabilistic to encourage diversity in language. 
\end{itemize}

Following~\cite{he2018emnlp}, the parser and the generator modules are obtained by rule-based method or supervised learning in advance, and fixed when training the dialog manager using supervised learning (SL) or fine turning using reinforcement learning (RL). 
The SL dialog manager employs a neural network to model state transitions $P(s_t|s_{t-1})$ (or equivalently, $\pi(a^i_t|s_{t-1})$) of the training corpus by minimizing the cross entropy loss. The RL dialog manager further fine tunes the SL model by maximizing a composed reward function with reinforcement learning.
The learned dialog policy $\pi(a^i_t|s_{t-1})$ can be further improved by enforcing some hand-craft rules.


There are two main problems with the SL or RL manager. 
First, the policy learned by an RL-based dialog manager produces reactive responses \cite{TamarLAWT16} , which are usually inadequate in a long term planning problem requiring more strategic thinking, such as negotiation. 
Second, it does not take the effect of the agent's generated utterances on opponents' reactions into account.
To address these problems, we propose an approach to incorporate the theory of mind (ToM) \cite{premack1978does} into the inference process. This enables one-step looking ahead to consider the effect of the agent's utterances and generate more thoughtful strategies.

\section{First-Order Theory of Mind for dialog}
\label{sec:method:tom}

The goal of the first-order theory of mind is to predict how a dialog act and an utterance generated by us would affect the reaction of the opponent. 
As illustrated in Figure \ref{fig:framework}, suppose that our current dialog state is $s_{t-1}$, which consists of the history of past dialog acts and the initial information, as well as the current utterance $u^{-i}_{t-1}$ from the opponent. The ToM model simulates the situations where we take dialog act $a^{i}_t$ ({\em e.g.}, {\tt propose(price=\$230)}) and utter the sentence $u^{i}_t$ ({\em ``how does \$ 230 for it sound"}), and estimates the probability distribution of the opponents response $a^i_{t+1}$.
By combining actions and states by definition, our first-order ToM model estimates the transition probability $T(s_{t+1} | u^{-i}_{t-1}, s_t, u^{i}_{t})$.

In practice, the opponent may have different
language preferences ({\em e.g.,} using more aggressive or mild words when countering) 
and strategies ({\em e.g.,} tend to insist on their target price or agree to a compromise). 
The first-order ToM can either {\bf implicitly} capture these personalities by learning the transition $T(s_{t+1} | u^{-i}_{t-1}, s_t, u^{i}_{t})$, or {\bf explicitly} infer the type of the opponent's personalities $z^{-i}$ first, from the past interaction and the opponent's utterance, i.e., learning an identifier $z^{-i}_{t-1} = f(s_{t-1}, u_{t-1}^{-i})$, and then learns the transition based on that information, i.e., $T(s_{t+1} | z^{-i}_{t-1}, s_t, u^{i}_{t})$, to make accurate prediction about opponents reaction.

\subsection{First-order ToM Policies with Explicit Personality Modeling} 
We introduce a policy with an explicit first-order ToM model $T(s_{t+1}|z^{-i} ,s_t, u^i_t)$, where the opponent's personality $z^{-i}$ can be estimated from partial dialog. During training, the ground truth of the type of opponents personalities, $z$, is given.  Therefore we can train an {\bf identifier} $z^{-i}_{t-1} = f(s_{t-1}, u^{-i}_{t-1})$ with extra supervision to predict the opponents type every round.
During the inference process, the probability of taking action $a^{i}_t$, i.e., a policy $\pi_{\tt ToM}(a^{i}_t | s_{t-1}, z^{-i}_{t-1})$, is proportional to 
\begin{equation*}
    \resizebox{0.48\textwidth}{!}{$\displaystyle
    \exp\left\{\frac{1}{\beta}\sum_{u^i_t}\underbrace{G(u^i_t|s_t, z_{t-1}^{-i})}_{\text{Generator}}\sum_{s_{t+1}} \underbrace{ T(s_{t+1}|z_{t-1}^{-i},s_t,u^i_t)}_{1^{\text{st}}\text{-order ToM}}\underbrace{V(s_{t+1})}_{\text{Value Fn.}}\right\},
    $}
\end{equation*}
where the exponent can be interpreted as the expected best return over opponent's next moves, after taking action $a_t^{i}$ at state $s_{t-1}$ (compressed as $s_t$). In the above expression, $T(s_{t+1}|z_{t-1}^{-i},s_t,u^i_t)$ is the {\bf explicit first-order ToM model}, which can be trained by supervised learning from the corpus; $G(u_t^{i}|s_t, z_{t-1}^{-i})$ is the {\bf generator} which renders utterance conditioned on the current state and the personality of the opponent; $V(s_{t+1})$, is the {\bf value function} estimated by the RL-based dialog manager, which gives the best future return estimation supposing the current state is $s_{t+1}$. It approximates $V(s_{t+1}, z_{t-1}^{-i})$  when it is nearly optimal. 
$\beta$ is the {\bf temperature parameter}. 
Since $\pi_{\tt ToM}$ is normalized as a Boltzmann distribution, when temperature $\beta \rightarrow \infty$, $\pi_{\tt ToM}$ is a uniform distribution over the next states;
when  $\beta \rightarrow 0$, $\pi_{\tt ToM}$ is nearly deterministic assigning most probability mass to the $s_t$ with the largest expected value after one-step ToM looking ahead. 

\subsection{First-Order ToM Policies with Implicit Personality Modeling}
\label{sec:method:implicit}

We also introduce first-order ToM policy with implicit personality modeling, 
where we do not have a module explicitly which explicitly predicts the opponent identity $z$. Instead, we combine the identifier and ToM model in the explicit version, to directly learn  $T(s_{t+1}|u_{t-1}^{-i}, s_t, u^i_t)$ without extra supervision.
In this case, $\pi_{\tt ToM}(a^i_t | s_{t-1}, u^{-i}_{t-1})$ is proportional to 
\begin{equation*}
    \resizebox{0.50\textwidth}{!}{$\displaystyle
    \exp\left\{\frac{1}{\beta}\sum_{u^i_t}\underbrace{G(u^i_t|s_t)}_{\text{Generator}}\sum_{s_{t+1}} \underbrace{ T(s_{t+1}|u_{t-1}^{-i},s_t,u^i_t)}_{1^{\text{st}}\text{-order ToM}}\underbrace{V(s_{t+1})}_{\text{Value Fn.}}\right\},
    $}
\end{equation*}
where $T(s_{t+1}|u_{t-1}^{-i}, s_t, u^i_t)$ is called the {\bf implicit first-order ToM model}, and the rest of components are similar to the explicit version.

We call $\pi_{\tt ToM}$ a first-order ToM policy, because it utilizes the first-order transition of the opponent, and estimates the expected outcome of performing a certain action which leads to state $s_t$. The personalities of the opponent are implicitly inferred from the previous utterance $u_{t-1}^{-i}$ and the history $s_t$. In practice, the summation (expectation) is approximated by Monte Carlo sampling.

\paragraph{Implicit vs Explicit model.}
We expect both explicit and implicit ToM models to provide several unique benefits. First, co-training the identifier $f(s_{t-1}, u^{-i}_{t-1})$ and the explicit first-order ToM model $T(s_{t+1}|z^{-i} ,s_t, u^i_t)$ is expected to have better sample efficiency than the implicit ToM model $T(s_{t+1}|u_{t-1}^{-i},s_t,u^i_t)$ since it utilizes the prior knowledge that personality identity affects state transition, and is trained with more supervision. Besides, with the personality $z^{-i}$, the generator and the value functions can also adapt to different populations of opponents.  However, the annotations for opponent types are not available for all corpora, therefore the implicit model would be a more general approach.

\subsection{Combining the RL Policy as a Prior}
\label{sec:method:combineRL}

After learning the above two ToM models from the corpus, we leverage the pre-trained RL policy as a prior with the 1st-order ToM policy to perform the {\bf inference}. The final policy is given by
    \vspace{-0.6em}
\begin{equation*}
    \resizebox{0.45\textwidth}{!}{$\displaystyle
    \pi(a^i_t|s_{t-1}, z_{t-1}^{-i}) \propto \pi_{\tt rl}(a^i_t|s_{t-1}) \cdot \pi_{\tt ToM}(a^i_{t} | s_{t-1}, z_{t-1}^{-i}),
    $}
    \vspace{-0.6em}
\end{equation*}
where $\pi_{\tt rl}$ is a policy obtained in a previous RL training process (see Section \ref{sec:neuroarch}). 

From a Bayesian point of view, $\pi_{\tt rl}$ can be seen as a prior $\mathbb P(a^i_t|s_{t-1})$, and the $\pi_{\tt ToM}$ is analog to the likelihood $\mathbb P(\text{best return}|a^i_t, s_{t-1})$ by its definition (not strictly true since it has to be summed up to one) which modifies the probability assignment in $\pi_{\tt rl}$, i.e., the posterior $\mathbb P(a^i_t|\text{best return}, s_{t-1})$. This gives the probability that the current agent should move to $s_t$ in order to reach the highest return in the end. 
$\pi_{\tt ToM}$ modifies the probability assignment in $\pi_{\tt rl}$, when $\beta \rightarrow \infty$ in $\pi_{\tt ToM}$, it is equivalent to the original RL policy $\pi_{\tt rl}$.

\begin{table*}[t]
    \centering
    {\small
    \begin{tabular}{lp{160pt}p{190pt}}
        \toprule
        Dialog Act & Definition & Example\\
        \midrule
        {\tt greet} & say hello or chat randomly. & ``Hello I am interested in buying."\\
        {\tt inquire} & ask any question about product, year, price, usage, etc. & ``How long have you had it? Does it come with any of the accessories?"\\
        {\tt inform} & provide information about the product, year, usage, etc. & ``It has the capability to offer great support for those over 6 foot."\\
        {\tt propose(price=)} & initiate a price or a price range for the product. & ``The list price is 600 but am willing to negotiate."\\
        {\tt counter(price=)} & propose a new price or a new price range (can be the same). & ``I'm sorry, I find homeopathy and any other pseudoscience to be profoundly upsetting as well. I'd be willing to go as high as \$5100."\\
        {\tt counter-noprice} & want to propose a new price but do specifically mention a new price. & ``I'm sorry, that's far too low. Since you know EC's reputation for quality, you know it's worth more than that."\\
        {\tt confirm} & ask question about with information to be confirmed. & ``Will the chair work for someone who is under 6 feet tall?"\\
        {\tt affirm} & give an affirmative response to a {\tt confirm}/{\tt propose}. & ``Yes absolutely the interface is quick and the phone is up to date as far the updates go."\\
        {\tt deny} & give a negative response to a {\tt confirm}/{\tt propose}. & ``No, I would expect that you would pick it up."\\
        {\tt agree(price=)} & make a deal. & ``Fair enough. \$30 it is!"\\
        {\tt disagree(price=)} & cannot make a deal. & ``I cannot take \$65 for something that is worth almost twice that.  Sorry, but no deal."\\
        {\tt offer(price=)} & final offer with price, no utterance & OFFER(\$65)\\
        {\tt accept} & final acceptance, no utterance & ACCEPT\\
        {\tt reject} & final rejection, no utterance & REJECT\\
        {\tt quit} & leave the negotiation, no utterance & QUIT\\
        \bottomrule
    \end{tabular}
    }
    \caption{Our redesigned dialog acts based on the {\sc CraigslistBargin} dataset, where {\tt propose}, {\tt counter}, {\tt agree}, {\tt disagree} are four intents must be followed by a {\tt price} slot, {\tt offer}, {\tt accept}, {\tt reject}, and {quit} are four terminal dialog acts with no corresponding natural language.}
    \label{tab:ontology}
\end{table*}

\section{Dialog Managers}
\label{sec:neuroarch}
We compare three hybrid {\bf dialog managers} combining neural networks and rules to control the flow of dialog:
\begin{itemize}
	\item [(1)] The {\bf SL+rule} manager employs a LSTM-based network to learn the transitions from $s_{t-1}$ to $s_{t}$ from corpus. Rules ensure that only deals meeting 70\% target are acceptable.
	\item [(2)] The {\bf RL} manager uses an actor-critic method \cite{mnih2016icml}, which contains a policy network with the same neural network architecture as the SL manager, and a value network predicts the future returns given states. 
	\item [(3)] The {\bf ToM} manager uses the first-order ToM policy as described in Section \ref{sec:method:tom}.
to learn the best response policy $\pi_{\tt ToM}(a^i_{t}|s_{t-1}, u_{t-1}^{-i})$ which is aware of the opponent's personalities and mental state. 


\end{itemize}

An extra LSTM model is used to encode $u^{-1}_{t-1}$ in both explicit and implicit ToM models, and learn the personality $z_{t-1}^{-i} = {\tt LSTM}(u^{-1}_{t-1}, s_{t-1})$ in explicit ToM models which encodes a distribution.
Note that for all three managers, we applied reasonable hand-crafted rules to prevent unreasonable policies. Specifically, the agent will never offer a price below its bottom line and will reject the opponent's offer if it is worse than its bottom line.

\paragraph{Training and Fine Tuning.} We first train the supervised learning ({\bf SL}) manager to minimize a loss function for the  dialog act predictions
\[
	\mathcal L_{SL} = {\tt CE}_{\text{intent}} + \alpha \cdot {\tt MSE}_{\text {price}},
\]
which is a linear combination of the cross entropy loss between the predicted intent and the ground truth intent, and the mean squared error between the predicted price and the ground truth price.
The reinforcement learning ({\bf RL}) manager is then fined tuned from the {\bf SL} manager to maximize a reward function described in Section \ref{sec:setup:reward}, with the actor-critic methods \cite{mnih2016icml}. The actor network is initialized as the {\bf SL} manager's LSTM-based network, and the critic network is partially initialized with the same network, followed by a MLP to predict the value.

For the {\bf ToM} manager, we reuse $V(s_{t+1})$ from a well trained {\bf RL} manager's critic network, and fix it during inference. 
The {\bf implicit first-order ToM model} $T(s_{t+1}|u_{t-1}^{-i},s_{t},u^{i}_t)$ is directly trained via supervised learning to minimize the same loss $\mathcal L_{SL}$.
For the {\bf explicit first-order ToM model}, $T(s_{t+1}|z_{t-1}^{-i},s_{t},u^{i}_t)$, we first train a LSTM-based identifier $z_{t-1}^{-i}=f(s_{t-1}, u^{-i}_{t-1})$, which receives ground truth opponent personality $z^{-i}$ from the corpus during training.  $T(s_{t+1}|z_{t-1}^{-i},s_{t},u^{i}_t)$ is learned with the input from the well-trained identifier.

To obtain the {\bf 1st-order ToM policy} for the inference, we approximate the sum (expectation) in $\pi_{\tt ToM}$ by Monte Carlo sampling with the generator, and discretize the price in a normalized price range. In practice, we found quantizing the price range with 100 units is a good balance between time comsumption and the quality of approximation.

\section{Experimental Setup}
\label{sec:setup}

We test our ToM negotiation framework on the {\sc CraigslistBargain} \cite{he2018emnlp}, which contains 6682 human-human dialogs between a buyer and a seller alternately bargaining for the price of an item on Craigslist. 

\paragraph{Ontology.}
We redesign the ontology of the {\sc CraigslistBargain} dataset to support a more diverse dialog act than the original coarse dialog acts \cite{he2018emnlp}, which can reflect more ways of mental state change in a negotiation.  We used the Microsoft Language Understanding Intelligent Service (LUIS) to relabel the dataset 
, and merged some similar label types,  such as {\tt insist} and {\tt vague-price} into {\tt counter-noprice}, and {\tt intro} and {\tt great} into {\tt greet}. All fifteen dialog acts after our modifications are in Table \ref{tab:ontology}. There are four intents {\tt propose}, {\tt counter}, {\tt agree}, {\tt disagree} that must be followed by a {\tt price} slot, and four terminal acts {\tt accept}, {\tt reject}, and {quit}. When an agent takes an {\tt offer} action, the other agent has to respond with {\tt accept} or {\tt reject}. Note that the function of this dialog act is not to capture the full semantic meaning of one utterance, but to serve as a logical skeleton for the dialog.

\paragraph{Reward function design.}
\label{sec:setup:reward}
We set the reward $r^{i}$ for the agent $i$ to be a linear function of the final price, such that the buyer achieves maximal reward of 1 at its target price, the seller achieves maximal reward of 1 at the listing price, and both agents receive zero rewards at the midpoint of the listing price and the target price. When there is no deal, both agents receives equivalent penalty.

\paragraph{Diverse opponent populations.}
All our negotiation experiments are conducted against variations of the {\bf SL+rule} manager as the opponent. For the variations, we create 7 different opponent populations (id=0$\sim$6) by injecting different rules for changing prices and rendering utterances. Price changing rules are functions of the number of sentences in the conversation history, which model the agreeability and the flexibility of a person. When rendering utterances, we use a template-based language generator as in 
\cite{he2018emnlp}, and insert population-specific tokens in utterances by sampling according to different opponent types.

The {\bf cooperative} population (id=5) will gradually compromise and move its price from the midpoint. The utterances of this population also contain more polite and mild words indicating its negotiable position. The {\bf most aggressive} population (id=0) will insist its price until the end, and utters more stubborn words. The {\bf competitive} population (id=6) compromises from target price slower than the {\bf cooperative}. The other populations will follow price changing curves in between these two extremes, 
and also have different language properties. The population types are accessible during training as ground truth values of $z_i$ to provide supervision (see Appendix \ref{app:setup} for details). 

\paragraph{Models.}
\label{sec:setup:design}

The {\bf dialog managers} we compare are described in Section \ref{sec:neuroarch}.
For the utterance {\bf parser}, we use Microsoft Language Understanding Intelligent Service (LUIS) \cite{williams2015sigdail} with 10 annotated training examples for each dialog act. For the {\bf Generator}, we use a retrieval-based model similar to \citealp{he2018emnlp} which samples an utterance from the top 10 matched templates.

\paragraph{Evaluation Metrics.}
We evaluate generated dialogs across four aspects: 
\begin{enumerate}
    \item {\bf Agreement rate} ({\tt Ag}), which is the percentage of dialogs that reach agreements.
    \item {\bf Objective utility} ({\tt Ut}), which is given by  \vspace{-4pt}
    $$\small \text{\tt Ut}^i = \begin{cases}
    	(P_{deal} - P_{target}^{-i})/\Delta P, & {\text{deal;}}\\
    	0, & {\text{no deal}}
    \end{cases} $$
    where $P_{deal}$ is the final deal price, the and total price range $\Delta P = P_{target}^{i} - P_{target}^{-i}$, where $P_{target}^{i}$, and $P_{target}^{-i}$ are the extreme target prices of the two agents. Note that this is different from the \textit{subjective utility} of each agent based on only its own price range, which may result in utilities $>1$ or $<0$ more often.
    \item {\bf Deal fairness} ({\tt Fa}), which is only for completed deals, as $\text{\tt Fa}^i = 1 - 2*\vert \text{\tt Ut}^i-0.5 \vert$.
    \item {\bf Dialog length} ({\tt Len}), which is the average turns of sample dialogs.
\end{enumerate} 

\begin{table*}[t]
    \centering
    
    \resizebox{\textwidth}{!}{
        \begin{tabular}{cccccccccccccc}
    	\toprule
		\multirow{2}{3em}{Method} &  \multicolumn{4}{c}{Cooperative Opponents (id=5)} 
		       & \multicolumn{4}{c}{Competitive Opponents (id=6)} 
		       & \multicolumn{5}{c}{Mixed Population (id=0$\sim$6)} \\
		\cmidrule(lr){2-5}\cmidrule(lr){6-9}\cmidrule(lr){10-14}
		{} & {\tt Ag $\uparrow$ } & {\tt Ut $\uparrow$ } & {\tt Fa $\uparrow$} & {\tt Len $\downarrow$}
		& {\tt Ag $\uparrow$ } & {\tt Ut $\uparrow$ } & {\tt Fa $\uparrow$} & {\tt Len $\downarrow$}
		& {\tt Ag $\uparrow$ } & {\tt Ut $\uparrow$ } & {\tt Fa $\uparrow$} & {\tt Len $\downarrow$} & {\tt Re $\uparrow$}\\
		\midrule
		SL+rule
		&$0.006$ & ${0.006}$ & $0.00$ & ${\bf  10.51}$
		&$0.005$ & ${0.005}$ & $0.00$ & ${\bf  10.64}$
		&$0.009$ & ${0.008}$ & $0.00$ & ${\bf  10.59}$ & ${-0.48}$
      	\\
		RL
		&$0.57$& $0.57$ & $0.00$ & $15.48$
		&$0.42$& $0.18$ & $0.32$ & $16.10$
		&$0.47$& $0.38$ & $0.00$ & $15.79$ & ${0.00}$
        \\
        \hline		
        ToM (implicit) 
		&${ 0.76}$&${0.72}$&$0.00$ &$13.14$
		&${ 0.34}$ & ${ 0.20}$ & ${0.45}$ & $14.26$ 
		&${ 0.48}$& ${ 0.44}$ & $0.00$ & $13.87$ & ${0.15}$
        \\
        ToM (explicit) 
		&${\bf 0.88}$&${\bf  0.78}$ & ${\bf  0.03}$ & $11.34$
		&${\bf 0.44}$ & ${\bf 0.24}$ & ${\bf  0.55}$ & $12.10$ 
		&${\bf 0.56}$& ${\bf 0.47}$ & ${\bf 0.10}$ & $11.74$ & ${\bf 0.16}$
        \\
		\bottomrule
    \end{tabular}
    }
    \vspace{-0.5em}
    \caption{
    Agreement rate ({\tt Ag}), agent utility ({\tt Ut}), deal fairness ({\tt Fa}), dialog length ({\tt Len}), and reward ({\tt Re})  for dialog managers (SL, RL, implicit and explicit ToM, with the best $\beta=0.05$) playing against cooperative, competitive, and mixed populations. Small negative fairness scores are truncated as zeros in the table. 
    }
    \label{tab:overall}
\end{table*}

\begin{table*}[t]
    {\small
    \begin{minipage}{.48\textwidth}
        \begin{tabular}{p{200pt}}
        \toprule
        \textit{Item: Phone, Listing price: \$50} \\
        \textbf{A: ToM(buyer), bottom: \$50, top \$38 }\\
        B: Cooperative(seller), bottom: \$35, top: \$50\\
        \midrule
        \textbf{A: Hello are you selling the blackberry?} \\
        B: Sure thing! What price are you thinking? \\
        \textbf{A: Absolutely, but the price is a little higher than I would like} \\
        B: My phone is in excellent condition and priced to sell quickly. \\
        \textbf{A: Would you be willing to take \$38.} \\
        B: Yea I can do \$44 for it. You come pick it up from me though. in person. Thanks \\
        \textbf{A: I think so, too! Sounds great!} \\
        B: OFFER(\$43)    \textbf{A: ACCEPT} \\
        \bottomrule
        \end{tabular}
    \end{minipage}
    \begin{minipage}{.04\textwidth}
    ~
    \end{minipage}
    \begin{minipage}{.48\textwidth}
        \begin{tabular}{p{205pt}}
        \toprule
        \textit{Item: Furniture, Listing price: \$50} \\
        \textbf{A: ToM(buyer), bottom: \$50, top: \$38} \\
        B: Competitive(seller), bottom: \$35, top: \$50 \\
        \midrule
        \textbf{A: Hi. How are you?} \\
        B: maybe. what do you have in mind? \\
        \textbf{A: How about \$38 dollars} \\
        B: Look, this is an expensive chair if you are willing to buy it somewhere else. I-m only selling it cause I'm moving out of the country.Otherwise I'd keep it but I don't want to just give it away, you know? How about \$49? \\
        \textbf{A: ok. i can do \$46 if you split the shipping in half} \\
        B: If you come pick it up today, that sounds fair. \\
        \textbf{A: Sounds fair to me, thank you} \\
        B: OFFER(\$48)    \textbf{A: ACCEPT} \\
        \bottomrule
        \end{tabular}
    \end{minipage}
    }
    \vspace{-0.5em}
    \caption{Example dialogs generated by ToM (explicit) against Cooperative (left) / Competitive (right) opponents. 
    }
    \label{tab:dialog_example}
    \vspace{-8pt}
\end{table*}
\section{Results}
\label{sec:results}

\paragraph{Improvement of dialog policy.} We evaluate {\bf SL+rule}, {\bf RL}, and our {\bf ToM} model on a mixed population for 4352 dialogs, which contains about 630 dialogs for each population. As shown in Table \ref{tab:overall}, 
our explicit {\bf ToM} model consistently achieves the highest agreement rate ({\tt Ag}), with 56\%, 4\%, and 20\% improvements compared to vanilla RL against cooperative, competitive, and mixed populations, respectively. Though deal agreement is hard for competitive opponents, our explicit {\bf ToM} model achieves more than 30\% improvement on the deal utility when interacting with this population. On the mixed population, the reward ({\tt Re}) for {\bf SL+rule} agent is low, as it is not directly optimized for better reward. {\bf RL} agent improves the {\tt Re} a lot compared with the {\bf SL+rule} baseline. However, both {\bf ToM} agents achieve better reward even when compared with {\bf RL} agent, which shows the advantage of strategic planning. Besides, unlike the {\bf SL+rule} only pursues high utility when there is a deal, but ends with every low {\tt Ag}, our {\bf ToM} models best balance both the agreement rate and agent utility of each dialog, and outperforms {\bf SL+rule} and {\bf RL} for all populations. 

\begin{figure}[t]
	\centering
	\includegraphics[width=0.95\columnwidth]{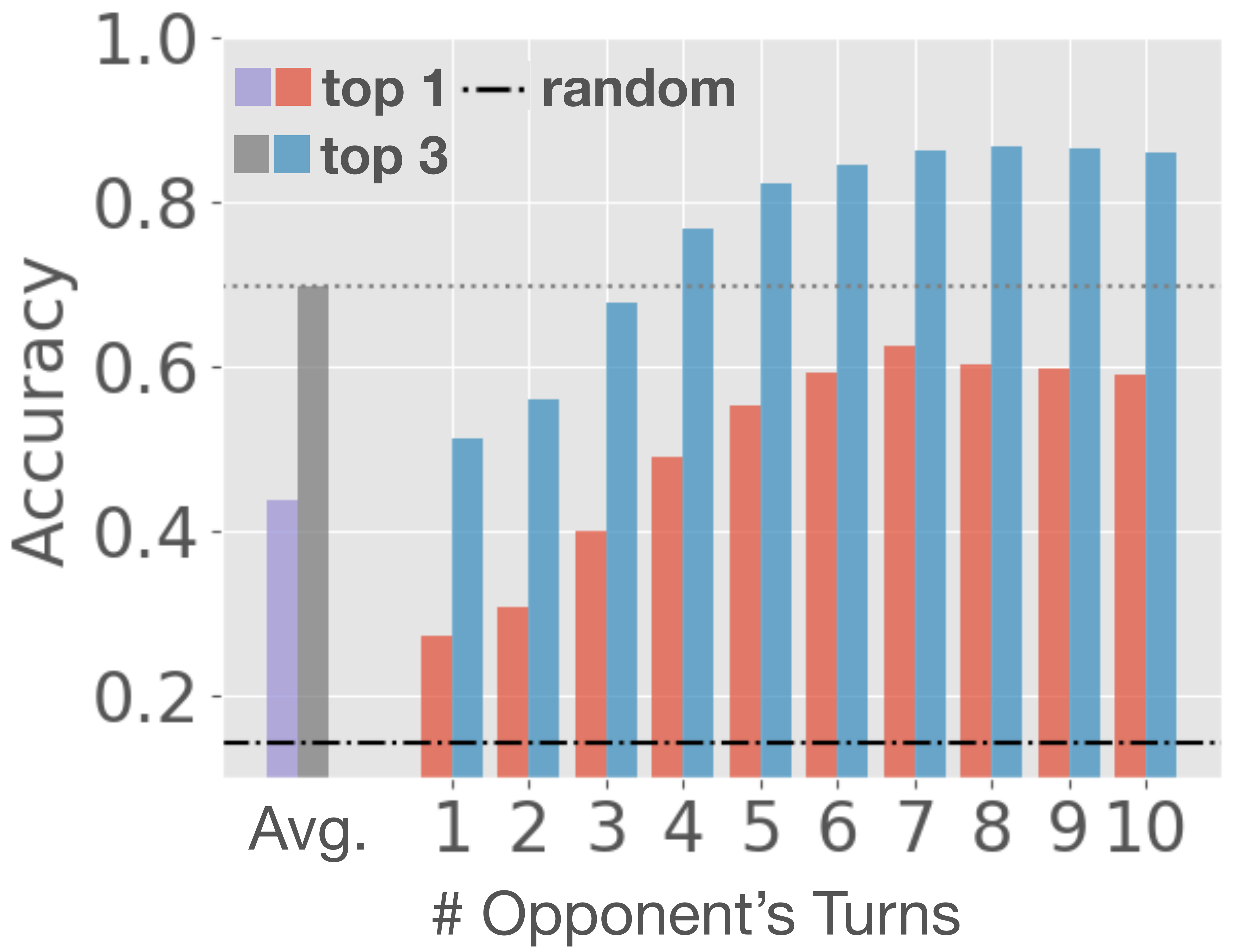}
	\vspace{-0.2em}
	\caption{Top 1 and top 3 accuracy of the characteristic identifier ($z$) during interaction with opponents.}
	\label{fig:identifier}
	\vspace{-1.2em}
\end{figure}

\paragraph{Implicit vs. explicit models.}
We found that the implicit \textbf{ToM} model can also achieve better {\tt Ag} and {\tt Ut} than the baselines for all populations. But the overall performance is slightly worse than the explicit {\bf ToM} model. This can be explained by the fact that the explicit model has more information about the population type during training. One may worry about the potential error cascade issue the explicit ToM models, as we see in Figure \ref{fig:identifier}, the top 1 accuracy of the identifier in the explicit model is only 69\%, though it is significantly above the chance. Our experiment show that even with an imperfect identifier, the explicit model can still outperform an implicit model, which is directly optimized for better performance.



\paragraph{Population-aware strategies.}
As Table \ref{tab:overall} shows, the {ToM} model can provide more deal fairness ({\tt Fa}, normalized price difference to the midpoint) to competitive opponents, since they rarely compromise, meanwhile reaching higher {\tt Ag} and {\tt Ut}. When opponents are cooperative and easy to negotiate with, our { ToM} model can achieve much better agent utility by taking advantage of losing some dialog fairness.
This implies our { ToM} model is able to utilize different characteristics of the opponents in the strategy generation.

We provide some sample dialogs from the explicit ToM model in Table \ref{tab:dialog_example}. When the seller is competitive, the buyer can adaptively raise its price and exchange for additional benefits, e.g., ``ok. i can do \$46 if you split the shipping in half" , to make the deal happen. We note that sometimes the offer prices slightly deviate from the agreed prince in negotiation but the ToM agent still accepts. This may be because the deflects of SL-based opponents is predictable to the ToM agent.

\begin{figure}[t]
	\centering
	\includegraphics[width=0.95\columnwidth]{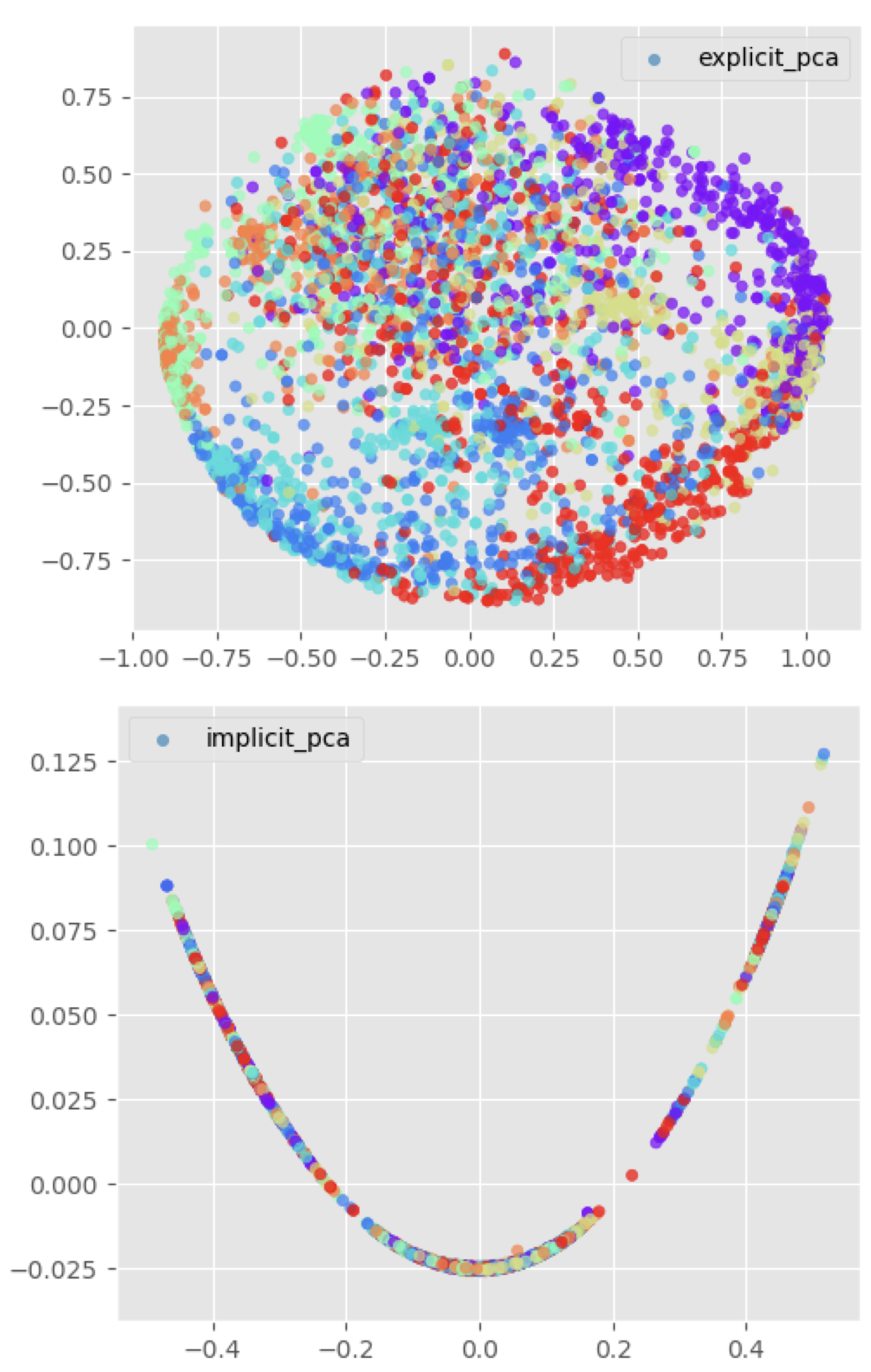}
	\vspace{-0.9em}
	\caption{PCA visualization of latent variables in the explicit (upper) and implicit (lower) ToM models. Colors indicate different opponent populations.}
	\label{fig:embedding}
\end{figure}

\paragraph{Effectiveness of the opponent identifier.} Figure \ref{fig:identifier} shows the identifier can capture the opponent identities well during interaction. The accuracy of the identifier increases as the dialog progresses. The top 1 accuracy after 6 opponent's turns is above 69\%, and the top 3 accuracy is above 84\%, where the chance is only 14.2\%. The average top 1 accuracy is 43.8\% for all turns in 5000 dialogs of different lengths. We also find the explicit ToM models can better prevent overfitting than implicit models. More details are in appendix \ref{app:result}.

\paragraph{Visualization of population embeddings.} In Figure \ref{fig:embedding}, we show the PCA visualization of the normalized latent variables in both explicit and implicit { ToM} models. The latent variables are extracted from one layer before the output of the identifier or its equivalence in the implicit model.  The explicit { ToM} model learns embeddings encoding different opponent populations, as the major variances of variable are captured by the difference of opponent populations. However, without extra supervision, the extraction of the population identity is difficult in the implicit ToM model. Further analysis shows that the variances of the latent variables in the implicit {ToM} model are mainly explained by intent types. We include more detailed analysis and t-SNE visualization in appendix \ref{app:result}.

\section{Conclusion}
In this work, we proposed a novel framework to integrate the concept of Theory of Mind (ToM) into generating  task-oriented dialogs. Our approach provides the ability to model and infer personality types of opponents, predict changes in their mental state, and use this information to adapt the agent's high-level strategy in negotiation tasks. We introduced a probabilistic formulation for first-order ToM and introduce two ways to incorporate it into a dialog agent, by 1) explicitly and 2) implicitly modeling the personality of the opponent. We tested our approach on a modified version of the {\sc CraigslistBargain} dataset~\cite{he2018emnlp} with diverse opponents. Our experiments show that our method using ToM inference achieves about {\bf 20\%} higher dialog agreement rate and utility compared to baselines on a mixed population of opponents. When negotiating with the cooperative opponents, the improvement of agreement rate is {\bf 54\%}. Some directions for future work include developing efficient schemes to approximate the value computation for future states, exploring higher orders of ToM, as well as a tighter integration of ToM into utterance generation and processing.
\section*{Ethical Considerations}
Our dataset is modified from the open-sourced {\sc CraigslistBargain} dataset~\cite{he2018emnlp}, which consists of negotiation dialogs between sellers and buyers on items from the Craigslist website. The initial dataset was collected using crowd workers on Amazon Mechanical Turk (AMT) playing the role of buyers and sellers. We redesigned the ontology to support more diverse dialog acts than the original coarse dialog acts. We manually labeled 10 examples for each intent, and used the Microsoft Language Understanding Intelligent Service to relabel the whole dataset. We create seven different populations by injecting different rules about changing prices and rendering utterances. 

Our paper involves an NLP application that can negotiate with people to reach agreement on deals. It is still at an early exploration stage so we do not expect it will currently cause any negative social impact such as massive job loss. If a mature version of such a system is deployed in the future, it may lead to less fair deals between the AI system and humans, as the system is optimized to find the best strategy that maximizes its own utility. But overall, we believe it will encourage market efficiency. 
\section*{Acknowledgements}
We thank Robert Hawkins, Jens Tuyls, Vishvak Murahari, Howard Chen and members of the Princeton NLP group for helpful discussions and feedback. This research was supported by an Amazon Research Award.

\bibliographystyle{acl_natbib}
\bibliography{acl2021}

\newpage
\appendix

~
\begin{center}
 {\Huge\bf Appendix}
\end{center}
~
\hrule\hrule\hrule

\section{Experimental Setup}
\label{app:setup}

To test our proposed framework in a realistic persuasive negotiation setting, we use the {\sc CraigslistBargain} dataset \cite{he2018emnlp}, which contains 6682 human-human dialogs between a buyer and a seller alternately bargaining for the price of an item on Craigslist. The listed price and a description is presented to both agents, and a private price is assigned to the buyer as the target. We set the reward $r^{i}$ to be a linear function of the final price, such that the buyer achieves maximal reward of 1 at its target price, the seller achieves maximal reward of 1 at the listing price, and both agents receive zero rewards at the midpoint of the listing price and the target price. When there is no deal, both agents receives equivalent penalty of -0.5.

\paragraph{Ontology}
\label{sec:setup:ontology}
We redesign the ontology of the {\sc CraigslistBargain} dataset to support a more diverse dialog act than the original coarse dialog acts \cite{he2018emnlp}, which can reflect more ways of mental state change in a negotiation. A dialog act consists of intent and a set of arguments. In our experiments, we only focus on the {\tt price} as it is the most important goal of this task. All fifteen dialog acts are listed in Table \ref{tab:ontology}. There are four intents {\tt propose}, {\tt counter}, {\tt agree}, {\tt disagree} that must be followed by a {\tt price} slot, and {\tt accept}, {\tt reject}, and {quit} are four terminal dialog acts with no utterance. When an agent takes an {\tt offer}, the other agent has to respond with {\tt accept} or {\tt reject}. Note that the function of this dialog act is not to capture the full semantic meaning of one utterance, but to serve as a logic skeleton of the dialog.

\paragraph{System Design}
\label{sec:setup:design}
{\bf Parser:} We use Microsoft Language Understanding Intelligent Service (LUIS) \cite{williams2015sigdail} with 10 starting training examples for each dialog act in our experiment. {\bf Generator:} We use a retrieval-based model similar to \citealp{he2018emnlp} which samples an utterance from the top 10 matched templates. 
We compared three hybrid {\bf dialog managers} combining neural nets and rules to control the flow of the dialog.
(1) The {\bf SL} manager employs a neural network to learn the transitions from $s_{t-1}$ to $s_{t}$ from dataset. We use a sequence model with two-layer LSTM with 300 hidden units for both the encoder and the decoder. (2) The {\bf RL} manager uses an actor-critic method \cite{mnih2016icml}, which contains a policy network with the same neural network architecture as the SL manager, and a value network predicts the cumulative reward given input states. The {\bf RL} manager also learns $\pi^i(s_{t}|s_{t-1})$ but with the goal of maximizing the total reward. (3) The {\bf ToM} manager uses the first-order ToM policy as described in \ref{sec:method:tom} to learn the best responding policy $\pi_{\tt ToM}(s_{t}|s_{t-1}, u_{t-1}^{-i})$ with the awareness of the opponent's characteristics and mental state change. An extra LSTM model is used to learn the characteristic identity $z_{t-1}^{-i}=f(s_{t-1}, u^{-i}_{t-1})$ in the explicit ToM model. For all three managers, we improve the learned policy by enforcing hand-craft rules. For example, the agent should never offer price below its bottom line and reject the opponent's offer if it is worse than its bottom line.

\paragraph{Populations of Opponents}
When playing against with a {\bf SL} manager, we create 7 different populations of opponents by injecting rules for changing the price and rendering utterance. 
Price changing rules are functions of the number of sentences in the conversation history, which model the agreeability and the flexibility of a person. The agreeability of a person is reflected in the range of relative prices (utility) at which a deal could be made. For example, a competitive opponent has a higher lower bound on the price, while a cooperative opponent has a lower initial price. The flexibility of a person is reflected in the slope and convexity of the price-changing rules. The price changing function for the most aggressive and stubborn opponent has a zero slope, encouraging them to  insist on their initial price until the end of the dialog. The more determined a seller is, the more concave the price changing function becomes. 

When rendering utterances, we use a template-based language generator as in \cite{he2018emnlp}, and insert population-specific tokens in utterances by sampling according to different opponent types. For example, in the utterances from a competitive opponent, words like ``afraid" or  ``unfortunately" appear more often, while words like ``great" or ``ok" will appear more frequently in the utterances from a cooperative opponent. Utterances of different populations should follow different distributions, and these sets of tokens are designed for this purpose.

We vary the price range, slope, and convexity to obtain the different behaviors for the seven different opponent types. 
The {\bf mildest} population will gradually compromise and lower its price (or raise its price if it is buyer). The utterances of this population also contain more polite and mild words indicating its negotiable position. And the {\bf most aggressive} population will insist its price until the end, and utters more stubborn words. The other five populations will follow different price changing curves in between these two extremes, and also have different language properties. All of these populations will deal at a certain price range, which depends on latest proposal price and current dialog length in different ways.

\paragraph{Training and Fine Tuning}
\label{sec:setup:training}

We train the {\bf SL} manager on 5000 dialogs for 20 epochs and choose the model with the lowest validation loss. The {\bf RL} manager is fine-tuned from a well-trained {\bf SL} agent by playing against itself. We choose the model with the highest reward.

For the {\bf ToM} manager, the value function $V(s_{t+1})$ is borrowed from a well trained {\bf RL} manager, and fixed during inference. The implicit {\bf first-order ToM model} $T(s_{t+1}|u_{t-1}^{-i},s_{t},u^{i}_t)$ is trained in a similar way as the {\bf SL} manager. To obtain the explicit first-order ToM model $T(s_{t+1}|z_{t-1}^{-i},s_{t},u^{i}_t)$, we co-trained it with a LSTM-based identifier $z_{t-1}^{-i}=f(s_{t-1}, u^{-i}_{t-1})$ for 2,000 episodes. 
Each run on training each manager was performed using a single NVIDIA GTX 2080 Ti GPU with 16GB RAM in approximate 2 hours. All the managers were trained using Adam with learning rate 0.001. For the {\bf ToM} manager, hyperparameter $\beta$ was randomly searched in range of ${0.05, 0.1, 1, 10}$, and the setting with the best results was $\beta = 0.05$.
\section{Additional Experimental Results}
\label{app:result}

\begin{figure}[h]
    \centering
    \includegraphics[width=0.85\columnwidth]{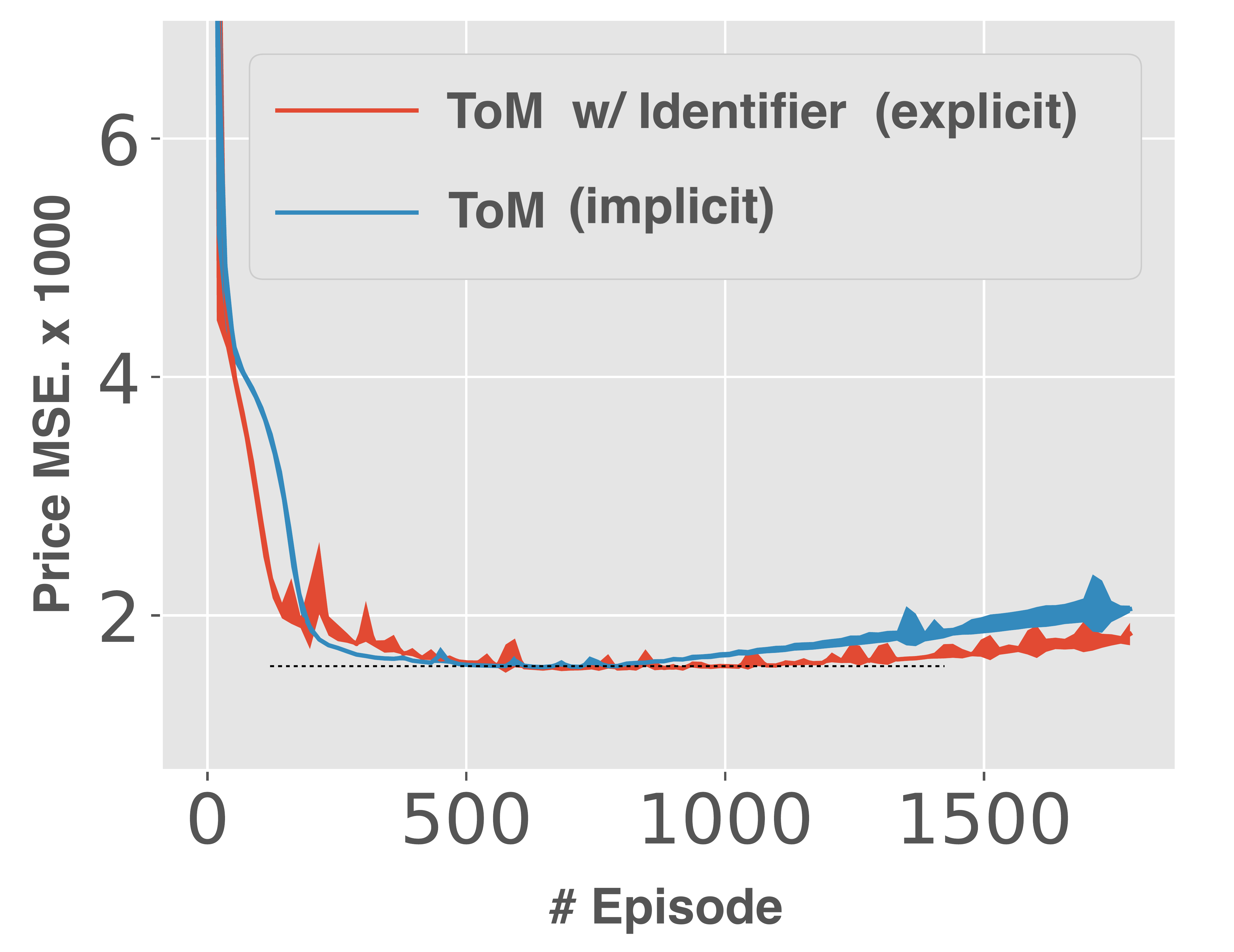}
    \vspace{-0.8em}
    \caption{The validation price MSE loss curves of implicit and explicit first-order ToM models. The explicit model with an identifier has slightly better sample efficiency and better prevents overfitting.}
    \label{fig:imp_vs_exp}
    \vspace{-2em}
\end{figure}

\subsection{Comparison of Implicit and Explicit ToM models}
We compared the implicit and the explicit ToM models as described in Appendix Section \ref{sec:method:tom} in the main article. Here additional Figure \ref{fig:imp_vs_exp} shows the validation mean squared error between the predicted price and the ground truth price of the opponent in the next turn (if exists) over 3,584 dialogs. Two models can both be trained well to perform this one-step prediction, while the explicit model with an identifier has slightly better sample efficiency and better prevents overfitting. This supports our hypothesis that the prior of different types of opponents is important.

\subsection{Visualization of Latent Variables}

\begin{figure}[h]
	\centering
	\includegraphics[width=0.75\columnwidth]{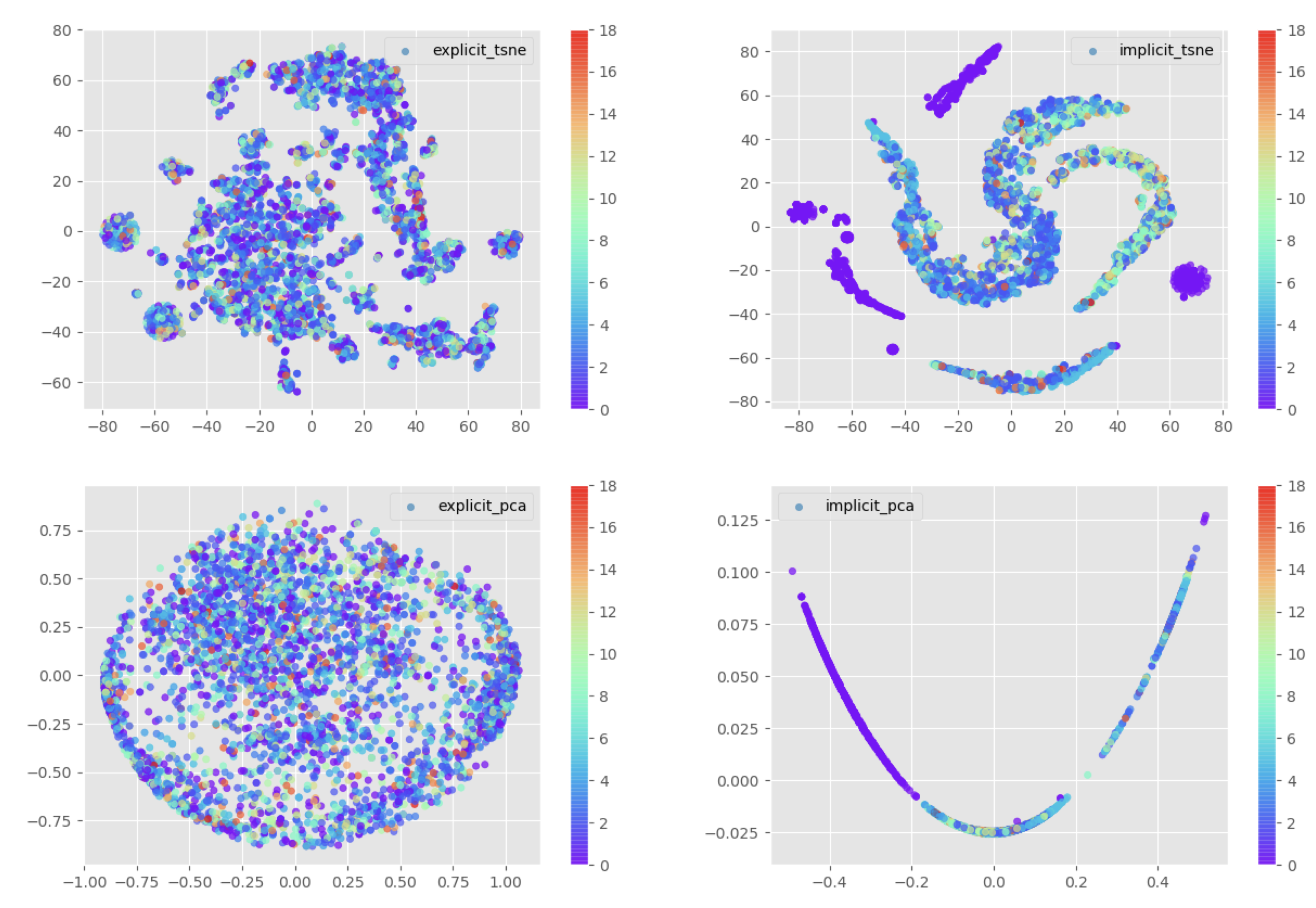}
	\vspace{-.5em}
	\caption{t-SNE and PCA visualization of latent variables in the explicit ToM model (left) and the implicit ToM model (right). Colors indicate different intents taken by the the agent.}
	\label{fig:embedding_my_action}
	\vspace{-1.2em}
\end{figure}

\begin{figure}[h]
	\centering
	\includegraphics[width=0.75\columnwidth]{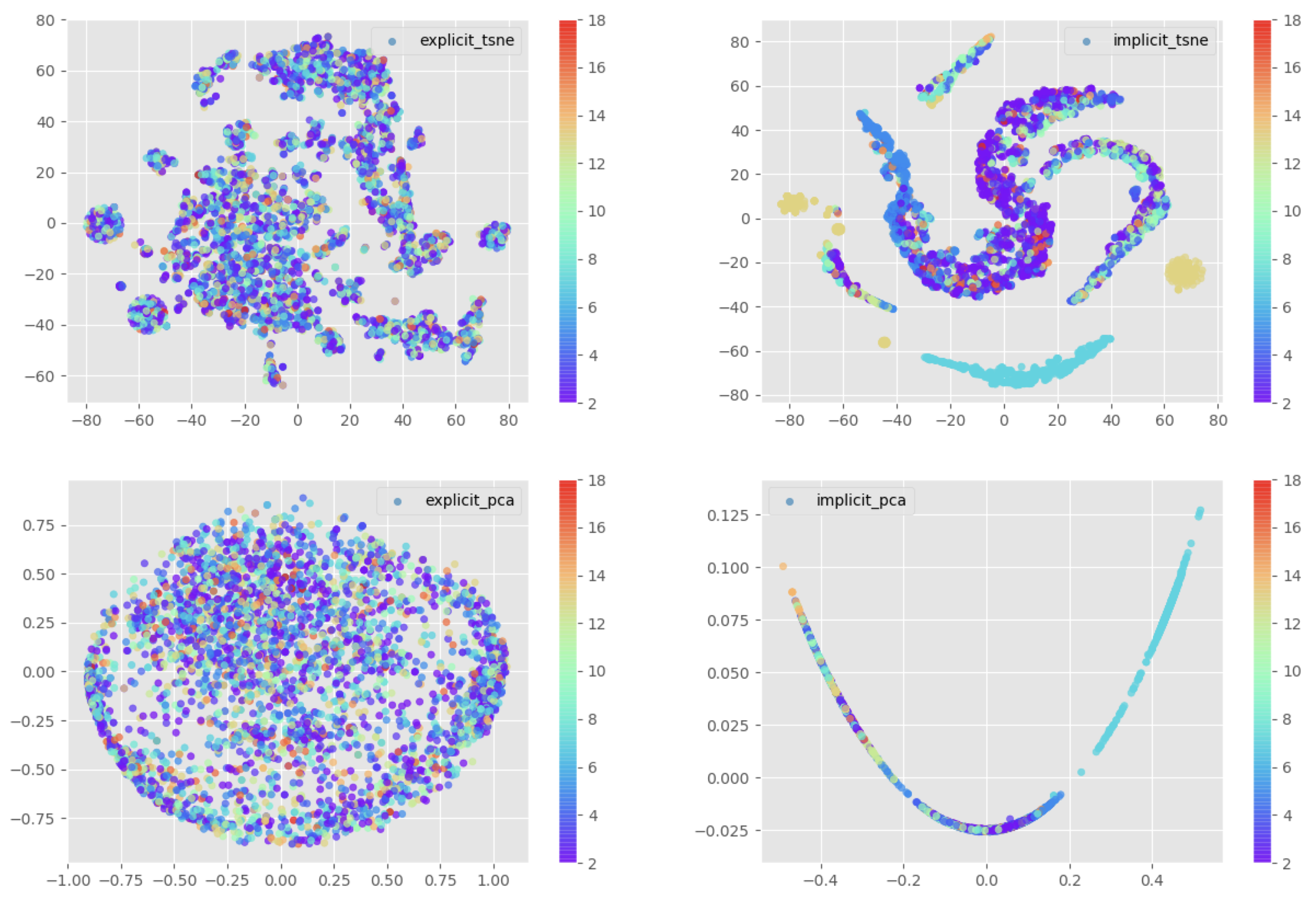}
	\vspace{-.5em}
	\caption{The same plot as Figure \ref{fig:embedding_my_action}. Colors indicate different intents taken by the the opponent.}
	\label{fig:embedding_oppo_action}
	\vspace{-1.2em}
\end{figure}

\begin{figure}[h]
	\centering
	\includegraphics[width=0.75\columnwidth]{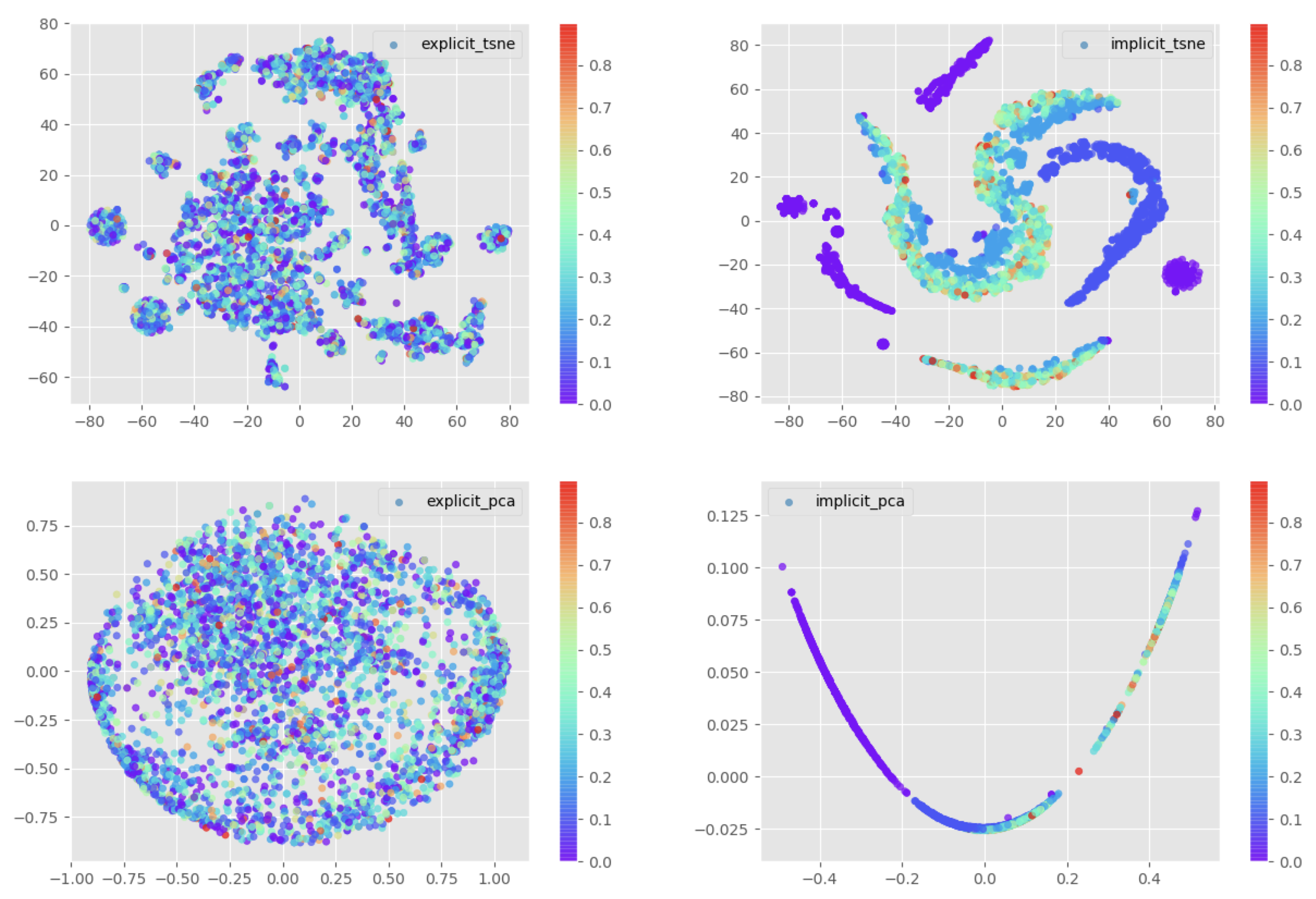}
	\vspace{-.5em}
	\caption{The same plot as Figure \ref{fig:embedding_my_action}. Colors indicate numbers of rounds of interaction.}
	\label{fig:turns}
	\vspace{-1.2em}
\end{figure}

\end{document}